\documentclass[journal]{IEEEtran}

\ifCLASSOPTIONcompsoc
  \usepackage[nocompress]{cite}
\else
  \usepackage{cite}
\fi

\ifCLASSINFOpdf
\else
\fi

\usepackage[utf8]{inputenc}
\usepackage{amsmath}
\usepackage{hyperref}
\usepackage{graphicx}
\usepackage[usenames, dvipsnames]{color}
\usepackage{amssymb}
\usepackage{bm}
\usepackage{algorithm}
\usepackage{algorithmic}
\usepackage{caption}
\usepackage{subcaption}
\usepackage{svg}
\usepackage{soul}
\usepackage{makecell}

\newcommand{\matr}[1]{#1}                 
\renewcommand{\vec}[1]{\mathbf{#1}}         
\newcommand{\unit}[1]{\hat{#1}}
\newcommand{\unitv}[1]{\mathbf{\unit{#1}}}  
\newcommand{\estm}[1]{\overline{#1}}       
\newcommand{\spce}[1]{\mathbb{#1}}

\newcommand{\T}{\mathsf{T}}  

\newcommand{\cmulti}[1]{\begin{tabular}[x]{@{}c@{}} #1 \end{tabular}}
\begin{document}
%
\title{Direct Triangulation with Spherical Projection \\ for Omnidirectional Cameras}

\author{Ciar{\'a}n~Eising
\thanks{C. Eising is with the Department
of Electronic and Computer Engineering, University of Limerick, Ireland.\protect\\
E-mail: ciaran.eising@ul.ie
}
}


\IEEEtitleabstractindextext{%
\begin{abstract}
In this paper, it is proposed to solve the problem of triangulation for calibrated omnidirectional cameras through the optimisation of ray-pairs on the projective sphere. The proposed solution boils down to finding the roots of a quadratic function, and as such is closed form, completely non-iterative and computationally inexpensive when compared to previous methods. In addition, even thought the motivation is clearly to solve the triangulation problem for omnidirectional cameras, it is demonstrated that the proposed methods can be applied to non-omnidirectional, narrow field-of-view cameras.
\end{abstract}

}

\maketitle

\vspace{1cm}

\IEEEdisplaynontitleabstractindextext

%
\IEEEpeerreviewmaketitle

\vspace{1cm}

\IEEEraisesectionheading{\section{Introduction}\label{sec:introduction}}

%
%
%
%
\IEEEPARstart{T}{riangulation} is the process of finding an estimate of a point in 3D space given its projection in at least two images and given the set of projection matrices that describe the relative position and orientations of the cameras (at least up to scale) and the camera intrinsics (except in the case of projective reconstruction). Naturally, in the absence of measurement uncertainty, this is a trivial, but in the presence of uncertainty, one should optimise the projected point. This is typically done for pinhole cameras without considering lens distortion. A very nice overview of the various methods of triangulation under the pin-hole assumption is described in \cite{Kukelova2019}, including discussions on linear triangulation, iterative techniques and $L_2$-norm optimal triangulation. As they provide a quite complete overview, readers are referred to that paper.

Omnidirectional cameras have found much use recently, as they have a distinct advantage of having extremely wide fields of view. Fisheye cameras have found useful applications over the last number of decades, including localisation \cite{gu2014fisheyelocalisation}, Simultaneous Localisation and Mapping \cite{Ji2020PanoramicSLAM}, and object detection \cite{zhu2018robotobjectdetection} among many others.  Such cameras offer a significantly wider field-of-view than standard cameras, often with greater than $180^\circ$ field-of-view. In consumer electronics, 360$^\circ$ cameras are becoming popular \cite{Jokela2019}, though the images from such camera are typically stitched images from a camera cluster, e.g., dual fisheye \cite{Tuan2017}.

In this paper, a completely closed form, non-iterative solution to the spherical triangulation problem is proposed, which selects the point that optimises the reprojection error to the unit sphere. Such an approach is suitable for omnidirectional cameras, as discussed in \cite{Forstner2016book}, as the spherical camera model can have a complete surround field of view. In fact, in \cite[\S13.4.1.1]{Forstner2016book}, F\"{o}rstner and Wrobel describe an iterative approach for triangulation based on the spherical camera model. What is proposed in this paper optimises the same function, and numerical equivalence to the F\"{o}rstner-Wrobel method is demonstrated. However, the triangulation proposed here is significantly simpler and computationally cheaper, as it directly solves the triangulation problem without iteration.

Of course, any two-view triangulation will always be suboptimal, due to the implicit assumption of complete knowledge of the relative pose. A rigorous solution is achieved only within a bundle adjustment framework, whereby the scene structure and camera pose are optimised over many frames. Triangulation is typically used to find initial values for bundle adjustment. However, one should not dismiss triangulation because of this, as a good initial seed is useful for the optimisation within bundle adjustment. Bundle adjustment is a topic for textbooks, and the reader is referred to \cite{Forstner2016book}.

\section{Spherical Triangulation}
\label{sec:triangulation}

In this section, the spherical triangulation methods are described. First, the spherical versions of the Midpoint and Direct Linear triangulation methods are presented, followed by two ray optimisation methods.

\subsection{Omnidirectional Camera Models} \label{sec::Fisheye}

What is initially required is an injective map from the image domain to the unit central projective sphere embedded in Euclidean space (itself embedded in $\spce{R}^3$)
$$g: {I}^2 \rightarrow {S}^2$$
where ${I}^2 \subset \spce{R}^2$ and ${S}^2 \subset \spce{R}^3$. In principle, any appropriate definition for the mapping function $\vec{u}~=~g(\vec{p}),\vec{u} \in {S}^2, \vec{p} \in {I}^2$ can be used. Such a mapping is done via a model of the camera geometry, of which several exist for omnidirectional cameras (e.g., as discussed in \cite{usenko2018doublesphere, hughes2010equidistant, khomutenko2016eucm}). 

\subsection{Euclidean and similarity reconstruction}

In the following sections, it is assumed that the camera poses, described by the rotations represented by the matrices $\matr{R}, \matr{R}' \in SO(3)$ and the camera positions represented by the vectors $\vec{C}, \vec{C}' \in \spce{R}^3$ are known in an external world coordinate system. This may not always be the case. Often the relative poses of the cameras are only known up to a undetermined scale factor. For two cameras, this is represented either as an essential matrix $\matr{E}$ or is decomposed to a rotation matrix $\matr{R}$ and a translation vector $\unitv{t}$ (recalling that $\matr{E} = R\left[\unitv{t}\right]_\times$, where $\left[\unitv{t}\right]_\times$ is the matrix representation of the cross product with $\unitv{t}$). The scale ambiguity is represented here by the fact that only a unit vector $\unitv{t}$ in the direction of translation is known, and the translation distance is unknown.

In such a case, a Euclidian triangulation (without scale ambiguity) is reduced to a similarity triangulation (with scale ambiguity). In the following sections, the position of the reference camera $\vec{C}$ can be replaced with the zero vector $\mathbf{0}$, the position of the second camera $\vec{C}'$ becomes $\unitv{t}$, the rotation of the first camera can be set to identity $\matr{I}$, and the rotation of the second camera is the relative rotation per the decomposition of the essential matrix. Beyond these adaptations, the following works with no issues for similarity triangulation.

\subsection{Midpoint method}
\label{sec:midpoint}

The midpoint method is outlined by \cite{hartley1997triangulation} for the case of point in the projective plane. Here, the equivalent method for points on the unit sphere is described. Assuming $\vec{u}$ and $\vec{u}'$ on the unit sphere are defined in a common coordinate system, the equation $\alpha\vec{u} - \alpha'\vec{u}' = \vec{C} - \vec{C}'$ is formed, giving three equations in two unknowns, which can be solved using linear least squares. In this case, $\alpha$ and $\alpha'$ define a point along the rays $\vec{u}$ and $\vec{u}'$ respectively that is closest to the other ray. The midpoint is given by $\estm{\vec{X}} = \left(\vec{C} + \alpha\unitv{u} + \vec{C}' + \alpha\unitv{u}'\right)/2$. Alternatively, to avoid an explicit least square algorithm, and where $\vec{t} = \vec{C}' - \vec{C}$ one can use:
\begin{align}
\alpha & = \frac{(\vec{u}'^\T \vec{u}')(\vec{u}^\T \vec{t}) - (\vec{u}^\T \vec{u}')(\vec{u}'^\T \vec{t})}{(\vec{u}^\T \vec{u})(\vec{u}'^\T \vec{u}') - (\vec{u}^\T \vec{u}')^2} \nonumber \\
\alpha' & = \frac{(\vec{u}^\T \vec{u}')(\vec{u}^\T \vec{t}) - (\vec{u}^\T \vec{u})(\vec{u}'^\T \vec{t})}{(\vec{u}^\T \vec{u})(\vec{u}'^\T \vec{u}') - (\vec{u}^\T \vec{u}')^2} \nonumber
\end{align}
In the result presented later, \textbf{Midpoint} is used to refer to this method.

\subsection{Spherical linear triangulation}
\label{sec:spherical_linear_triangulation}

Linear triangulation in the case of features on the projective plane is well known \cite{hartley1997triangulation}. However, consider $\vec{u}$ as the projection of $\vec{X}$ on to the unit sphere $S^2$. $\vec{u}$ is colinear with $\vec{X}$ in the absence of uncertainty. That is, $\vec{u} \times \left(\matr{R}\left(\vec{X} - \vec{C}\right)\right)~=~\vec{0}$, where $\matr{R} \in SO(3)$ and $\vec{C} \in \spce{R}^3$ describe the orientation and position of the camera respectively. The cross product can be made explicit as as
\begin{align}
    u_1 \vec{r}^{3\T} \left(\vec{X} - \vec{C}\right) - u_3 \vec{r}^{1\T} \left(\vec{X} - \vec{C}\right) & = 0 \nonumber \\
    u_2 \vec{r}^{3\T} \left(\vec{X} - \vec{C}\right) - u_3 \vec{r}^{2\T} \left(\vec{X} - \vec{C}\right) & = 0 \nonumber \\
    u_1 \vec{r}^{2\T} \left(\vec{X} - \vec{C}\right) - u_2 \vec{r}^{1\T} \left(\vec{X} - \vec{C}\right) & = 0  \nonumber
\end{align}
where $\vec{r}^{i\T}$ is the $i$th row of $\matr{R}$. An equation of the form $\matr{A}\vec{X} = \vec{b}$ can be formed, which can be solved using the pseudo-inverse $\estm{\vec{X}} = \matr{A}^\dagger\vec{b}$. In this case,
\begin{equation} \label{eqn:sphLinRecA}
    \matr{A} = \left(\begin{matrix} 
    u_1\mathbf{r}^{3\T}   - u_3\mathbf{r}^{1\T} \\ 
    u_2\mathbf{r}^{3\T}   - u_3\mathbf{r}^{2\T} \\ 
    u'_1\mathbf{r}'^{3\T} - u'_3\mathbf{r}'^{1\T} \\ 
    u'_2\mathbf{r}'^{3\T} - u'_3\mathbf{r}'^{2\T} \\
    \vdots
    \end{matrix}\right)
\end{equation}
and
\begin{equation} \label{eqn:sphLinRecb}
    \vec{b} = \left(\begin{matrix} 
    \left(u_1\vec{r}^{3\T} - u_3\vec{r}^{1\T}\right)\vec{C} \\ 
    \left(u_2\vec{r}^{3\T} - u_3\vec{r}^{2\T}\right)\vec{C} \\ 
    \left(u'_1\vec{r}'^{3\T} - u_3'\vec{r}'^{1\T}\right)\vec{C}' \\ 
    \left(u'_2\vec{r}'^{3\T} - u_3'\vec{r}'^{2\T}\right)\vec{C}' \\
    \vdots
    \end{matrix}\right)
\end{equation}
Naturally, at least two observations of $\vec{X}$ are required for linear triangulation. Multiple observations can be used for triangulation by extending in (\ref{eqn:sphLinRecA}) and (\ref{eqn:sphLinRecb}). In the result presented later, the term \textbf{Sph-Lin} refers to this method.

\subsection{Direct spherical triangulation}

\begin{figure}[!t]
  \centering
  \includegraphics[width=\linewidth]{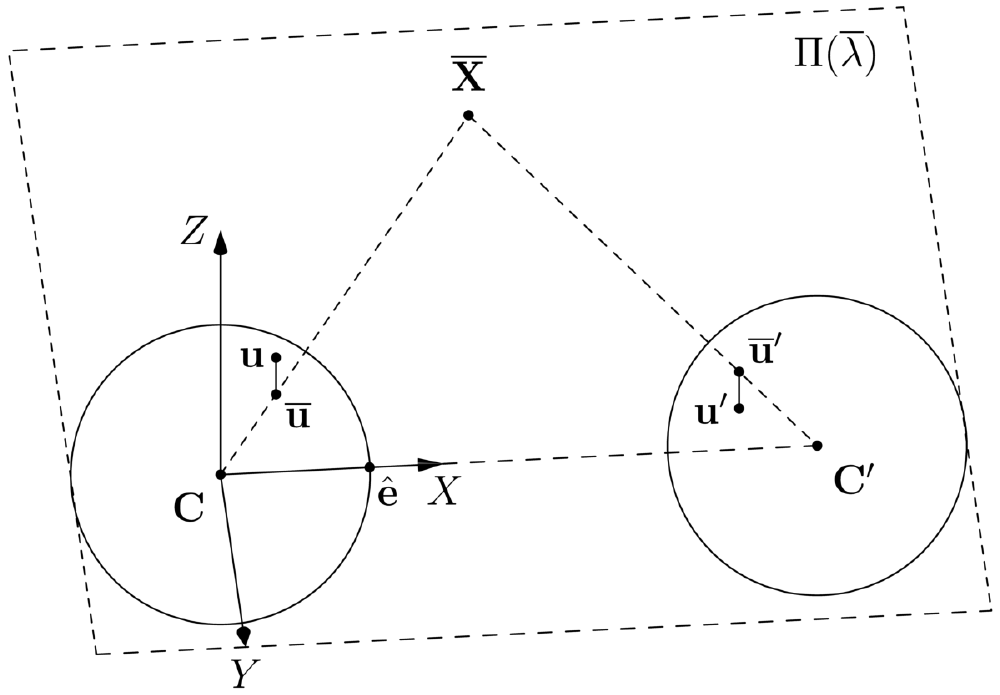}
  \caption{Spherical triangulation}
  \label{fig:sphrecon}
  \vspace{-3mm}
\end{figure}

\subsubsection{Cost function}

It has long been known that two points in a perspective image (denoted by the correspondence $\vec{p} \leftrightarrow \vec{p}'$) that observe the same 3D point, must adhere to the epipolar constraint in the absence of uncertainty, i.e. $\vec{p}'^\T \matr{E} \vec{p} = \vec{p}'^\T \left[\vec{t}\right]_\times\!\! \matr{R} \vec{p} = 0$ \cite{longuethiggins1981}. Geometrically, this means that the points should lie on the corresponding epipolar lines in each of the perspective images. Hartley and Sturm \cite{hartley1997triangulation} exploited this fact to find the optimal epipolar line pair in the presence of uncertainty, onto which the point correspondences were projected. However, this cannot be used for omnidirectional imagery, as the perspective image is a very poor model of such a camera. Instead, a similar process to select the optimal epipolar plane is employed.

In spherical coordinates, it is understood that a noisy correspondence pair $\vec{u} \leftrightarrow \vec{u}'$ does not generally satisfy the constraint $\vec{u}^\T \matr{E} \vec{u}'$. The goal is to find the pair of spherical coordinates $\estm{\vec{u}} \leftrightarrow \estm{\vec{u}}'$ that are as close to the original pair $\vec{u} \leftrightarrow \vec{u}'$, but satisfying the epipolar constraint, i.e., that minimise the function
\begin{equation} \label{eq:cost_funt_orig}
  \mathcal{C}(\vec{u},\vec{u}') = d(\vec{u},\estm{\vec{u}})^2 + d(\mathbf{u}',\estm{\vec{u}}')^2 \; \textrm{subject to} \; \estm{\vec{u}}'^\T \matr{E} \estm{\vec{u}} = 0
\end{equation}
This is similar at first glance to the cost function used by Hartley and Sturm \cite{hartley1997triangulation}, but in this case, $d$ refers 3D Euclidean distance (Figure \ref{fig:sphrecon}). The restriction that $\estm{\vec{u}}'^\T \matr{E} \estm{\vec{u}}~=~0$ means that the estimated pair $\estm{\vec{u}} \leftrightarrow \estm{\vec{u}}'$ must both lie on $\Pi$. The goal is, therefore, to find the optimal parameterisation of $\Pi$. This is also the same function that is optimised by the F\"{o}rstner-Wrobel triangulation \cite[\S13.4.1]{Forstner2016book}. However, as mentioned previously, F\"{o}rstner and Wrobel describe an iterative solution to finding the point optimised pair, whereas here a direct solution is proposed.

\subsubsection{Parameterising the plane}

The following relates to Figure \ref{fig:sphrecon}. In the case that an estimate of the essential matrix is available, $\matr{E}$ is decomposed into a rotation matrix $\matr{R}$ and translation vector $\mathbf{\hat{t}}$, which describe the relative pose of the two cameras. As the translation is resolved only up to scale, the translation vector $\mathbf{\hat{t}}$ is represented as a unit vector. $\mathbf{\hat{t}}$ defines the camera baseline and is in fact the epipole on the first spherical surface ($-\matr{R}^{-1}\mathbf{\hat{t}}$ defines the epipole on the second spherical surface).
The goal is to find the optimal epipolar plane, which is one of the pencil of planes defined by the baseline between the camera positions.

Firstly, observe that in ${S}^2$, the epipole $\unitv{e}$ is the image of the other camera position on the unit sphere, and lies on the baseline between the cameras (defined by $\unitv{t}$, given $\vec{t} = \vec{C}' - \vec{C}$). 
Now parameterise a pencil of planes in Hessian normal form:
\begin{equation} \label{eq:sheaf}
  \Pi(\lambda): \left(\vec{n}_1 + \lambda\vec{n}_2\right)^\T\vec{x} = 0
\end{equation}
$\vec{n}_1$ and $\vec{n}_2$ can be any non-parallel normal vectors perpendicular to $\unitv{e}$. It therefore makes some sense to pick normal vectors that simplify the problem as best possible. The basis of the first spherical image can be changed such that the $X$-axis aligns with the epipole $\unitv{e}$, by applying the appropriate rotation $\matr{R}_b$ to the measured feature $\vec{u}$. Note that $\matr{R}_b$ is fixed per camera pair, so can be applied to all features as a pre-processing.

If the $X$-axis is not already aligned with $\unitv{e}$, a rotation about the vector defined can be by 
\begin{align} \label{eqn:axisangle}
\vec{a} & = \left(1, 0, 0\right)^\T \times \unitv{e} \nonumber \\
\theta & = \arcsin{\left(\|\vec{a}\|\right)}
\end{align}
where $\vec{a}$ and $\theta$ are the axis-angle representation of the rotation and can be converted trivially to the matrix representation $\matr{R}_b$. $\|\cdot\|$ refers to the Euclidean norm. In the case that the $X$-axis is already aligned with $\unitv{e}$, then set $\matr{R}_b = \matr{I}$. From now on, feature $\vec{u}$ refers to the feature in the new rotated coordinate system. That is, all features $\vec{u}$ are replaced with $\matr{R}_b\vec{u}$. 

Now with $\unitv{e} = (1,0,0)^\T$, it is possible to select $\vec{n}_1 = (0,1,0)^\T$ and $\vec{n}_2 = (0,0,1)^\T$. This combination can be used to represent any plane that intersects the camera baseline, except for the plane orthogonal to $\vec{n_2}$, which is dealt with shortly. The parameterisation of the plane becomes:
\begin{equation} \label{eq:simpSheaf}
  \Pi(\lambda): \left(\left(\begin{matrix} 0 \\ 1 \\ 0 \end{matrix}\right) + \lambda\left(\begin{matrix} 0 \\ 0 \\ 1 \end{matrix}\right)\right)^\T\vec{x} = \left(\begin{matrix} 0 \\ 1 \\ \lambda \end{matrix}\right)^\T\vec{x} = \vec{n}^\T\vec{x} = 0
\end{equation}
where $\vec{n} = (0,1,\lambda)^\T$.

Here, $\lambda$ can be arbitrarily large. For example, in the case that $\vec{u}$ is close to the plane orthogonal to $\vec{n}_2 = (0,0,1)^\T$ (i.e., the $XY$-plane), $\lambda$ will be some large value, and as it approaches the $XY$-plane, numerical issues can dominate. It is, therefore, desirable to keep the value of $\lambda$ within some reasonable range. In the case that $\vec{u}$ is closer to the $XY$-plane than the $XZ$-plane (i.e., $|u_2| \geq |u_3|$), then set $\vec{n}_1 = (0,0,1)^\T$ and $\vec{n}_2 = (0,1,0)^\T$, i.e., $\vec{n}_1$ and $\vec{n}_2$ can be swapped. (\ref{eq:simpSheaf}) becomes
\begin{equation} \label{eq:simpSheaf_2}
  \Pi(\lambda): \left(\left(\begin{matrix} 0 \\ 0 \\ 1 \end{matrix}\right) + \lambda\left(\begin{matrix} 0 \\ 1 \\ 0 \end{matrix}\right)\right)^\T\vec{x} = \left(\begin{matrix} 0 \\ \lambda  \\ 1 \end{matrix}\right)^\T\vec{x} = \vec{n}^\T\vec{x} = 0
\end{equation}
where $\vec{n} = (0,1,\lambda)^\T$.
In all practical cases, by selecting between (\ref{eq:simpSheaf}) and (\ref{eq:simpSheaf_2}) based on the proximity of $\vec{u}$ to the $XY$- and $XZ$-planes, the valid range of $\lambda$ is kept to approximately $[-1,1]$. It may drift slightly above 1 or below -1 in some cases, depending on the second observation $\vec{u}'$.

\subsubsection{Minimizing sum of squares}

The aim is to pick the value of $\lambda$ that minimises the square distance of the observed correspondence pair $\vec{u} \leftrightarrow \vec{u}'$ to the plane $\Pi$. The cost function is therefore
\begin{equation} \label{eq:cost}
  \mathcal{C}(\mathbf{u},\mathbf{u}',\Pi(\lambda)) = d(\mathbf{u},\Pi(\lambda))^2 + d(\mathbf{u}',\Pi(\lambda))^2
\end{equation}
where $d$ denotes the Euclidean distance of the point on the unit sphere $\mathbf{u}$ to a given plane $\Pi$ with normal $\mathbf{\hat{n}}$. 

Again, to simplify a bit, put $\mathbf{u}'$ in the same orientation as $\mathbf{u}$, by applying the rotation between the camera orientations. If there exists an external world coordinate system, then $\matr{R}_r = \matr{R}\matr{R}'^\T$, where $\matr{R}$ and $\matr{R}'$ describe the orientation of the two cameras. Therefor $\vec{u}' := \matr{R}_b\matr{R}_r\vec{u}'$. From here, simply refer to this updated feature as $\mathbf{u}'$. 

The Euclidean distance of a point $\mathbf{u}$ to a plane $\Pi(\lambda)$ is given by
\begin{equation} \label{eq:distance}
  d(\mathbf{u},\Pi)^2 = \left(\frac{\vec{u}^\T\vec{n}}{\|\vec{n}\|}\right)^2
\end{equation}
and the total squared distance is given by
\begin{align} \label{eq:sqr_dist}
  s(\lambda) & = \frac{\left(\vec{u}^\T\vec{n}\right)^2 + \left(\vec{u}'^\T\vec{n}\right)^2}{\|\vec{n}\|^2} \nonumber \\
        &= \frac{a + \lambda b  + \lambda^2 c}{1 + \lambda^2}
\end{align}
where 
\begin{align}
a &= 
\begin{cases} 
u_3^2 + u_3^{\prime 2}, & |u_2| > |u_3| \\
u_2^2 + u_2^{\prime 2}, & \textrm{otherwise}
\end{cases}  \nonumber \\
c &= 
\begin{cases} 
u_2^2 + u_2^{\prime 2}, & |u_2| > |u_3| \\
u_3^2 + u_3^{\prime 2}, & \textrm{otherwise}
\end{cases} \nonumber \\
b &= 2(u_2u_3 + u'_2u'_3)
\end{align}
Elementary calculus tells us that the minima and maxima of $s(\lambda)$ occur when $s'(\lambda) = 0$. The derivative of $s(\lambda)$ is given by
\begin{equation} \label{eqn:deriv}
s'(\lambda) = -\frac{b\lambda^2 + 2(a - c)\lambda - b}{(1 + \lambda^2)^2}
\end{equation}
The solutions of $s'(\lambda) = 0$ occur when the numerator (which is a quadratic polynomial in $\lambda$) is equal to zero. Therefore, the numerator is equated with zero and the roots are obtained using the standard quadratic formula. In fact, the smaller of the two roots need only to be evaluated, as the smaller root will lead to a smaller value of (\ref{eq:sqr_dist}), i.e., the minima of $s(\lambda)$. Therefore, $\lambda$ can be directly estimated as:
\begin{equation} \label{eqn:quad_formula}
\estm{\lambda} = \frac{(c-a) - \sqrt{(a - c)^2 + b^2}}{b}
\end{equation}
$\estm{\lambda}$ will always be real, as $(a - c)^2 + b^2 \geq 0$. One must handle the case that $b=0$, in which case one must find the root of $(a - c)\lambda = 0$, i.e., $\estm{\lambda} = 0$, except in the case that $a=c$. In the case that $b=0$ and $a=c$, the derivative $s'(\lambda) = 0$, and thus there is no solution and $\estm{\lambda}$ remains undefined. Thus, the optimal plane normal is 
\begin{equation} \label{eqn:normalvec}
\estm{\vec{n}} = 
\begin{cases}
\left(0, \estm{\lambda}, 1\right)^\T, & |u_2| > |u_3| \\
\left(0, 1, \estm{\lambda}\right)^\T, & \textrm{otherwise}
\end{cases}
\end{equation}
The correspondences on the unit spheres $\vec{u} \leftrightarrow \vec{u}'$ for two views are corrected through orthogonal projection to the optimal plane $\Pi\left(\estm{\lambda}\right)$ described by the normal vector $\estm{\unitv{n}}$.
\begin{equation} \label{eqn:prProjectPlane}
\estm{\vec{u}} =
      \vec{u} - \frac{\vec{u}^\T\estm{\vec{n}}}{\|\estm{\vec{n}}\|^2}\estm{\vec{n}}
\end{equation}
This vector need not lie on the unit sphere. Of course, if needed, it can be projected to the unit sphere $\left(\frac{\estm{\vec{u}}}{\|\estm{\vec{u}}\|}\right)$. However, as is discussed soon, the \textbf{Midpoint} method will be used for the final triangulation, and this does not require the vectors to lie on the unit sphere. In the results presented later, this method is referred to as \textbf{Sph-Quad}.

\subsubsection{Minimizing sum of magnitudes}

Hartley and Sturm also proposed an approach to minimize the sum of magnitudes of distances for perspective images and demonstrated that this had slightly better performance than the sum of squares approach when considering reprojection error \cite{hartley1997triangulation}, albeit at the cost of requiring a root finding over an 8th order polynomial. It is worth investigating if a similar approach can be taken for the case of a spherical camera model. For the moment, assume that $|u_3| \geq |u_2|$, and as such $\vec{n}~=~\left(0,1,\lambda\right)^\T$. The aim is to select the value of $\lambda$ that minimizes
\begin{align} \label{eq:abs_dist}
  s_2(\lambda) 
             & = \frac{\left|u_2 + \lambda u_3\right| + \left|u'_2 + \lambda u'_3\right|}{\sqrt{1 + \lambda^2}} \nonumber \\
             & = \frac{\left(\omega u_2 + \omega' u_2'\right) + \lambda\left(\omega u_3 + \omega' u'_3\right)}{\sqrt{1 + \lambda^2}}
\end{align}
where $\omega$ and $\omega'$ are equal to -1 or +1 depending on the signs of $u_2 + \lambda u_3$ and $u'_2 + \lambda u'_3$ respectively.

The derivative is given as
\begin{equation} \label{eq:abs_dist_der}
  s_2'(\lambda) = \frac{\left(\omega u_3 + \omega' u'_3\right) - \lambda\left(\omega u_2 + \omega' u_2'\right)}{\left(1 + \lambda^2\right)^{\frac{3}{2}}}
\end{equation}
Setting the numerator equal to zero and rearranging gives
\begin{equation}
 \omega\left(u_3 - u_2 \lambda \right) = \omega'\left(u_2' \lambda - u'_3\right) 
\end{equation}
$\omega$ and $\omega'$ are functions of $\lambda$, the value of which is being sought. It is desirable to remove these, so square both sides, and rearrange to give a quadratic in $\lambda$.
\begin{equation}
(u_2^2 - u_2'^2)\lambda^2 + 2(u_2'u_3' - u_2u_3)\lambda + (u_3^2 - u_3'^2) = 0
\end{equation}
The roots of this are found using the quadratic formula, resulting in
\begin{equation} \label{eqn:abs_dist_quad}
\estm{\lambda} = \frac{(u_2u_3 - u_2'u_3') \pm (u_2'u_3 - u_2u_3')}{u_2^2 - u_2'^2}
\end{equation}
The correct root is found by selecting the value of $\lambda$ that returns the smallest value of (\ref{eq:abs_dist}).

Following the above steps for the case that $|u_2| > |u_3|$ and $\vec{n}~=~\left(0,\lambda,1\right)^\T$, it can be seen that only the denominator of (\ref{eqn:abs_dist_quad}) changes to $u_3^2 - u_3'^2$. Thus, the complete solution for $\estm{\lambda}$ is
\begin{align}
\estm{\lambda} & = \frac{(u_2u_3 - u_2'u_3') \pm (u_2'u_3 - u_2u_3')}{d} \nonumber \\
d & = \begin{cases} 
u_3^2 - u_3'^2, & |u_2| > |u_3| \\
u_2^2 - u_2'^2, & \textrm{otherwise}
\end{cases} 
\end{align} 
In the result presented later, this method is referred to as \textbf{Sph-Abs}.


\subsubsection{Note on the geodesic distance to epipolar plane}

It is perhaps a little bit more natural to think about the error on the surface of $S^2$, as this is the projection surface, in place of the Euclidean error in $\mathbb{R}^3$. 
The geodesic distance is the angle between the point $\mathbf{u}$ and the great circle defined by the intersection of the plane $\Pi$ with the unit sphere $S^2$:
\begin{align} \label{eq:geodesic}
  s(\lambda) & = \left(\arccos\left(\frac{\mathbf{u}^\T\mathbf{n}}{\|\vec{n}\|}\right) - \frac{\pi}{2}\right)^2 + \left(\arccos\left(\frac{\mathbf{u}'^\T\mathbf{n}}{\|\vec{n}\|}\right) - \frac{\pi}{2}\right)^2 \nonumber\\
  & = (\alpha)^2 + (\alpha')^2 \nonumber
\end{align}
where $\alpha$ and $\alpha'$ are the angles between the plane $\Pi$ and $\mathbf{u}$ and $\mathbf{u}'$ respectively. Using the small angle assumption ($\alpha \approx 0$):
\begin{align}
  s(\lambda) & \approx \sin^2(\alpha) + \sin^2(\alpha') \nonumber \\
  s(\lambda) & \approx \left(\frac{\mathbf{u}^\T\mathbf{n}}{\|\vec{n}\|}\right)^2 + \left(\frac{\mathbf{u}'^\T\cdot\mathbf{n}}{\|\vec{n}\|}\right)^2
\end{align}
 Noting that the $\alpha^2 \approx \sin^2(\alpha)$ for small values of $\alpha$, then the same solution as (\ref{eq:sqr_dist}) is realised. 

The same can be realised for the minimization of the magnitudes.
\begin{align} \label{eq:geodesic}
  s_2(\lambda) & = |\alpha | + |\alpha'| \approx |\sin(\alpha)| + |\sin(\alpha')| \nonumber \\
  s_2(\lambda) & \approx \left|\frac{\mathbf{u}\cdot\mathbf{n}}{\|\vec{n}\|}\right| + \left|\frac{\mathbf{u}'\cdot\mathbf{n}}{\|\vec{n}\|}\right|
\end{align}
which is the same solution as in (\ref{eq:abs_dist}). The assumption that the angle is small ($\alpha \approx 0$) is valid; a large angle would mean a poor feature registration, or a feature on a moving object, which cannot be triangulated.

\subsubsection{Triangulation}

The pair $\estm{\vec{u}} \leftrightarrow \estm{\vec{u}}'$ are now on $\Pi\left(\estm{\lambda}\right)$, and, except for numerical issues, will intersect at a unique point in $\spce{R}^3$. A basis for the subspace of $\spce{R}^3$ that is analogous to the plane $\Pi\left(\estm{\lambda}\right)$ could be found, and the intersection in the subspace calculated. The linear method defined by (\ref{eqn:sphLinRecA}) and (\ref{eqn:sphLinRecb}) could be used. However, a computationally less expensive approach, and equally valid, is to use the \textbf{Midpoint} method discussed earlier.

F\"{o}rstner and Wrobel \cite[Alg.~21]{Forstner2016book} describe a spherical triangulation method that can return a point at infinity and handle rays that converge in a backward direction. Their triangulation method is based on the observation that the view of the end points of a line segment in one camera is the mathematical equivalent of the observation of a single point from two camera locations, with the camera baseline playing the role of the line segment \cite[\S13.3.3,~\S12.3.3]{Forstner2016book}. Without going into the detail (for which the reader is referred to the mentioned book), it can be thought of as measuring the parallelity of the vectors $\estm{\vec{u}}$ and $\estm{\vec{u}}'$.

\subsection{Assumptions of the proposed solution}

The proposed method is optimal if it is assumed that the uncertainty in the ray directions is isotropic and homogeneous. The former indicates that the covariance matrix is $\sigma^2 \vec{I}$, with the norm constraint enforced. The latter indicates that variance $\sigma$ of the correspondence pair $\vec{u} \leftrightarrow \vec{u}'$ are equal. Additionally, the sum-of-squares optimisation is optimal in the case of Gaussian uncertainty, whereas the sum-of-absolutes solution is optimal if in the case of Laplacian uncertainty. A great deal of space is dedicated in \cite[\S10.2, \S12.2.1]{Forstner2016book} to discussing uncertain camera coordinates, both for planar and spherically normalised coordinates. While not wishing to replicate what is contained in that work, it is pertinent to make some observations. 

All uncertainty models are approximations of the actual uncertainty characteristics. Consider a non-omnidirectional camera, that is one whose image is reasonably close to perspective, but with radial distortion. Most methods of triangulation in this case assume isotropic and homogeneous coordinate uncertainty on the corrected Euclidean plane \cite{hartley1997triangulation, Kanatani2008}. In the presence of radial distortion, the standard approach is to first undistort the coordinates. This, however, means that the uncertainty model assumed by those methods is no longer correct. The image undistortion alters the uncertainty characteristics. Kukelova and Larsson \cite{Kukelova2019}, it seems, are the first to attempt to optimise the projection of the point in the presence of radial distortion, providing an iterative method that is based on either the single-parameter division model \cite{fitzgibbon2001divisionmodel} or the two-parameter polynomial \cite{brown1966} models of radial distortion. While the method of \cite{Kukelova2019} can be extended to higher order versions of the division or polynomial distortion models, it is not easy to see if it is extendable to omnidirectional models, that often incorporate trigonometric functions, or have no analytic inverse in the case of polynomial projection models.

Regardless of the applicability to omnidirectional cameras, the work of Kukelova and Larsson only addresses radial distortion. There are, however, other sources of inhomogeneity and anisotropy of point uncertainty. For example, as discussed in \cite[\S13.3.1]{Forstner2016book}, if correlation is used for correspondence, the structure tensor scaled with image uncertainty provides an approximation of the uncertainty of the correspondence displacement. Optics can impact the performance of point correspondences, as lens characteristics tend to worsen towards the image periphery, particularly for omnidirectional/fisheye cameras \cite{Schneider2016}. Artefacts such as vignetting and reduced optical resolution incur a deviation from the assumed uncertainty model. The point spread function of a camera, particularly fisheye cameras, are often far from isotropic \cite{Lehmann2019}.

The aim, of course, is not to dismiss triangulation methods as fruitless. Quite to the contrary, it would seem that despite these confounding factors, triangulation performance in work over the last couple of decades is at least acceptable for the given applications, particularly when acknowledging that, in fact, all triangulation methods are suboptimal as they assume a known and uncertainty-free relative pose of the camera(s), which is solved only in the context of bundle adjustment.

\section{Experimental Evaluation}

\label{sec:results}

In the results presented, \textbf{F-W} refers to the F\"{o}rstner-Wrobel Triangulation method \cite[Alg.~21]{Forstner2016book}, and \textbf{Pln-Poly} refers to Hartley-Sturm's polynomial optimisation on the image plane \cite{hartley1997triangulation}, along with the terms already introduced (being, \textbf{Midpoint}, \textbf{Sph-Lin}, \textbf{Sph-Quad}, and \textbf{Sph-Abs}).

In this section, an experimental evaluation of the methods in this paper, compared to existing methods, is presented. The aims are threefold: to demonstrate 1) the numerical equivalence to \textbf{F-W}, 2) the effectiveness of the method for both omnidirectional and non-omnidirectional cameras, and 3) the computational efficiency compared to existing methods. To compare spherical methods, an omnidirectional set of \textit{near points} is created. For non-omnidirectional reconstruction, a set of narrow field-of-view \textit{far points} is used. The following outcomes of the experiments can be expected:
\begin{itemize}
\item \textbf{Sph-Quad} and \textbf{F-W} should return numerically the same results across all tests, as they optimise the same function.
\item In the case that Gaussian uncertainty is added to the synthetic data, \textbf{Sph-Quad} should perform better than \textbf{Sph-Abs}. The inverse should be true for when Laplacian uncertainty is added.
\item All of the methods that optimise the ray pair should perform better than the methods that do not (being \textbf{Midpoint} and \textbf{Sph-Lin}), particularly for the reprojection error.
\item \textbf{Sph-Quad} and \textbf{Sph-Abs} should be computationally faster than the other optimisation methods, as they are direct methods with no iterative parts.
\item For non-omnidirectional cameras, \textbf{Pln-Poly} should outperform the spherical methods. However, this should be mild, considering that the surface of the sphere approaches the planar surface as the field-of-view shrinks.
\end{itemize}

\subsection{Evaluation on Synthetic Data}

\subsubsection{Synthetic data creation}

\begin{figure}
     \centering
     \begin{subfigure}[b]{0.25\linewidth}
         \centering
         \includegraphics[height=2.3cm]
         {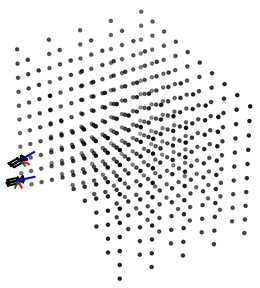}
     \end{subfigure}
     \begin{subfigure}[b]{0.33\linewidth}
         \centering
         \includegraphics[trim=32 20 0 0,clip,height=2.3cm] {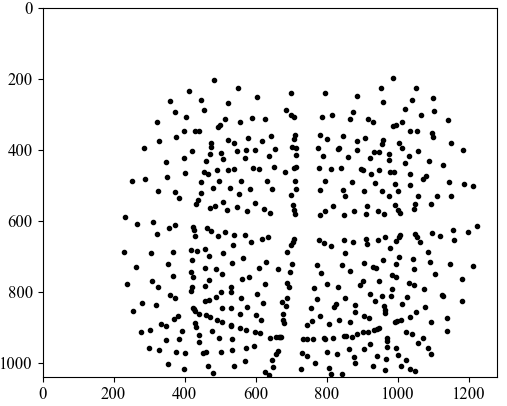}
     \end{subfigure}
     \begin{subfigure}[b]{0.33\linewidth}
         \centering
         \includegraphics[trim=34 19 0 0,clip,height=2.3cm]
         {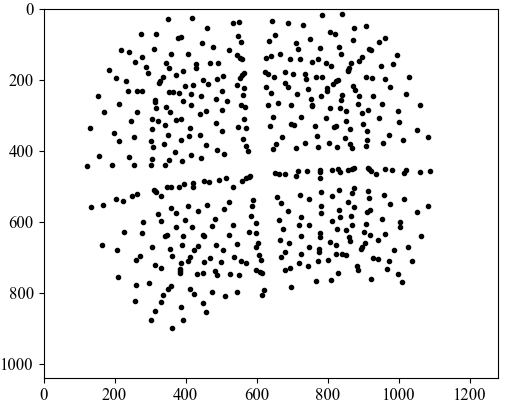}
     \end{subfigure}
    \caption{Synthetically generated test data, showing the grid of 3D points (\textit{near points}) and the randomised camera positions and the two synthetically created images.}
    \label{fig:3D_gt}
    \vspace{-3mm}
\end{figure}

The triangulation methods are tested on synthetically generated 3D data. The scenes were generated by defining a grid in 3 dimensions that is 20 units in the X and Z dimensions, and 10 units in Y dimensions. For the omnidirectional set of points, the grid of points is placed 1 unit from the primary camera (\textit{near points}), and for the narrow field-of-view set of points, it is place at 10m from the camera (\textit{far points}). The primary camera is fixed at the origin, and the second camera is randomly placed at unit distance. A random rotation to the second camera about each axis is applied in the range $(-10, 10)$ degrees. In the case of the tests on the spherical triangulation, this rotation is inconsequential. Figure \ref{fig:3D_gt} shows the grid of features in $\spce{R}^3$.

For the case of modelling uncertainty on the sphere, the points are projected to each of the unit spheres, giving the true coordinates on the spherical image pair $\widetilde{\vec{u}} \leftrightarrow \widetilde{\vec{u}}'$. The direction vectors are perturbed giving the modelled measurement $\vec{u} \leftrightarrow \vec{u}'$. The perturbation is added by applying a random rotation to the vectors $\vec{u}$ and $\vec{u}'$, the rotation being represented as the quaternion $\left(1, a/2, b/2, c/2\right)^\T$ \cite[\S8.1.5.3]{Forstner2016book}, where $a$, $b$ and $c$ are random normal variables (mean 0, variance $\sigma$). When modelling uncertainty on the omnidirectional image or on the projective plane, a 2D random normal variable is added to the projected location.

Finally, the performance of the triangulation when the uncertainty is in the omnidirectional image is tested. A fisheye camera is modeled by projecting the 3D points to the image using the double sphere model. The parameters of the BF2M2020S23 lens ($195^\circ$ field-of-view) from \cite{usenko2018doublesphere} are used. The parameter set for this model and camera is $\{f_x~=~313.21$, $f_y~=~313.21$, $c_x~=~638.66$, $c_y~=~514.39$, $\xi~=~-0.18$, $\alpha~=~0.59\}$. The image size is $1280 \times 1040$ pixels. The camera is only calibrated using data from within the image area, and as such the model may not be valid outside of this range. Therefore, any points from the set that do not project to within these pixel limits are removed. Figure \ref{fig:3D_gt} shows the two omnidirectional images of the grid of features.

\subsubsection{Comparison to F\"{o}rstner-Wrobel Triangulation}

As mentioned previously, both \textbf{Sph-Quad} and \textbf{F-W} optimise the same function under the coplanarity constraint (\ref{eq:cost_funt_orig}). Given that both methods optimise the same function, they should return the same results, within some tolerances to allow for computational differences (for example, the fact that the\textbf{F-W} is iterative with a finite number of iterations). It is worth demonstrating this computationally to remove any doubts.

\textbf{Sph-Quad} and \textbf{F-W} are applied to the synthetic spherical image data \textit{near points} in the presence of Gaussian uncertainty, returning the optimised pairs $\estm{\vec{u}}_\textrm{sq} \leftrightarrow \estm{\vec{u}}'_\textrm{sq}$ and $\estm{\vec{u}}_\textrm{fw} \leftrightarrow \estm{\vec{u}}'_\textrm{fw}$ respectively. In the \textbf{F-W}, a convergence tolerance of $T=0.01$ is used. In Table \ref{tab:forstcomp}, the mean and max differences of the estimates and the mean residuals are shown, for varying levels of uncertainty. The difference between the vectors is measured as the $L_2$-norm, i.e., $d = \|\estm{\vec{u}}_\textrm{sq} - \estm{\vec{u}}_\textrm{fw}\|$ and $d' = \|\estm{\vec{u}}'_\textrm{sq} - \estm{\vec{u}}'_\textrm{fw}\|$. The residual is $r_\textrm{sq} = \|\estm{\vec{u}}_\textrm{sq} - \vec{u}\| + \|\estm{\vec{u}}'_\textrm{sq} - \vec{u}'\|$, and similarly for the case of \textbf{F-W} triangulation. The difference between the resultant optimised vectors is very small, and in particular the residuals are almost identical. 

\begin{table}[tbp]
\caption{Comparison to F\"{o}rstner-Wrobel \cite[Alg.~21]{Forstner2016book}}
\vspace{-0.3cm}
\begin{center}
\begin{tabular}{c c c c c}
\hline
\vspace{-0.2cm} \\
& \multicolumn{3}{c}{Added Gaussian $s^2$ uncertainty ($\sigma$)} \\
                                                                              & 0.001                     & 0.01                      & 0.1  \\
\hline \vspace{-0.2cm} \\
 $\textrm{mean}(d)$   & $6.2221\!\times\!10^{-7}$ & $4.7716\!\times\!10^{-5}$ & $2.8092\!\times\!10^{-3}$ \\
 $\textrm{mean}(d')$ & $7.2358\!\times\!10^{-7}$ & $4.8320\!\times\!10^{-5}$ & $2.7893\!\times\!10^{-3}$ \\
 $\max(d)$            & $2.5331\!\times\!10^{-4}$ & $6.8885\!\times\!10^{-4}$ & $3.7750\!\times\!10^{-2}$ \\
 $\max(d')$          & $3.8477\!\times\!10^{-4}$ & $7.0994\!\times\!10^{-4}$ & $3.4351\!\times\!10^{-2}$ \\
 $\textrm{mean}(r_\textrm{fw})$                  & $2.1573\!\times\!10^{-3}$ & $2.1634\!\times\!10^{-2}$ & $2.1736\!\times\!10^{-1}$ \\
 $\textrm{mean}(r_\textrm{sq})$                  & $2.1572\!\times\!10^{-3}$ & $2.1634\!\times\!10^{-2}$ & $2.1737\!\times\!10^{-1}$ \\
\vspace{-0.2cm} \\
\hline
\end{tabular}
\label{tab:real_values}
\end{center}
\vspace{-0.2cm}
\label{tab:forstcomp}
\vspace{-3mm}
\end{table}

\subsubsection{Synthetic Results}

\begin{figure*}
    \centering
    \begin{subfigure}[b]{0.32\textwidth}
    \centering
        \includegraphics[height=4.3cm]{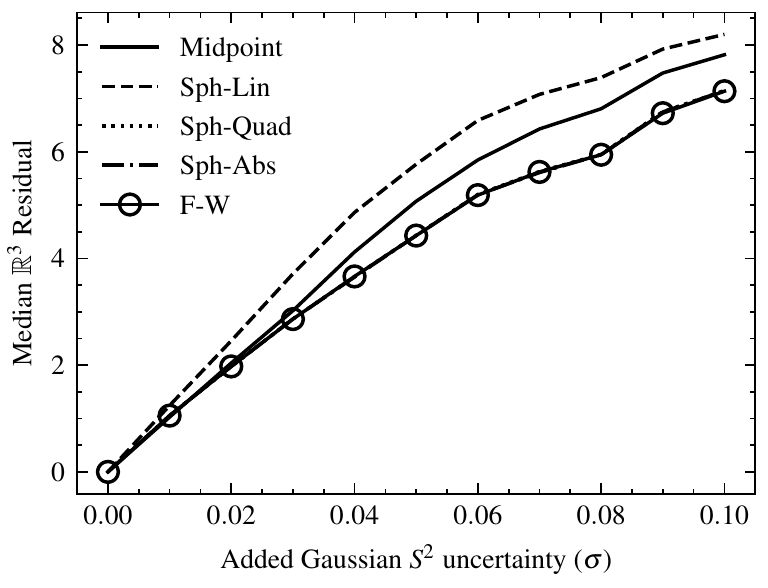}
    \end{subfigure}
    \begin{subfigure}[b]{0.32\textwidth}
    \centering
        \includegraphics[height=4.3cm]{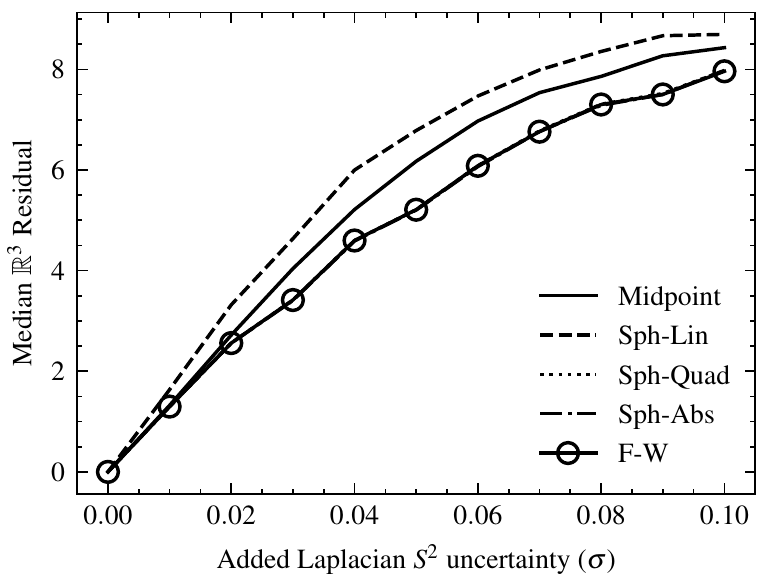}
    \end{subfigure}
    \begin{subfigure}[b]{0.32\textwidth}
    \centering
        \includegraphics[height=4.3cm]{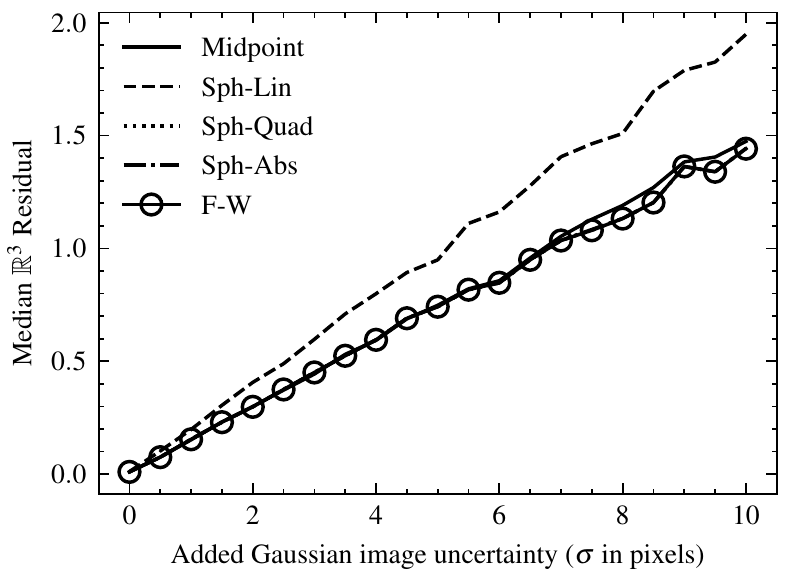}
    \end{subfigure}
    \begin{subfigure}[b]{0.32\textwidth}
    \centering
        \includegraphics[height=4.3cm]{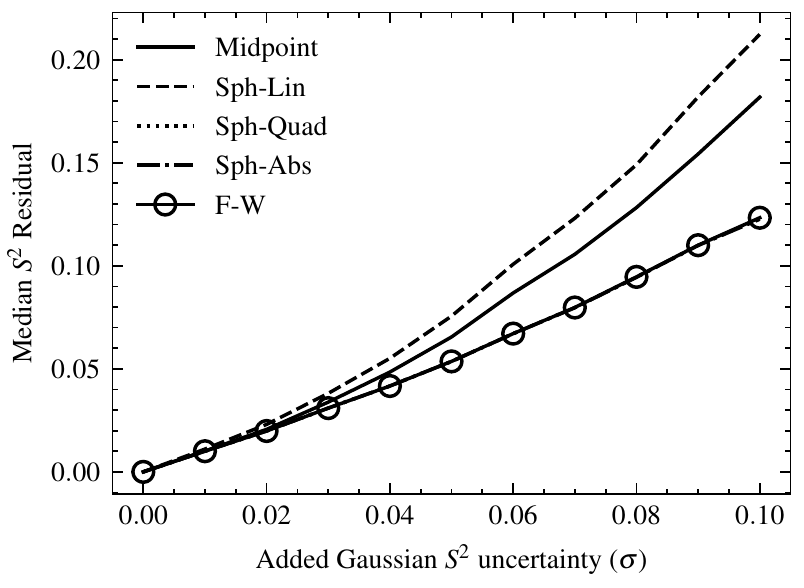}
    \end{subfigure} \hspace{0mm}
    \begin{subfigure}[b]{0.32\textwidth}
    \centering
        \includegraphics[height=4.3cm]{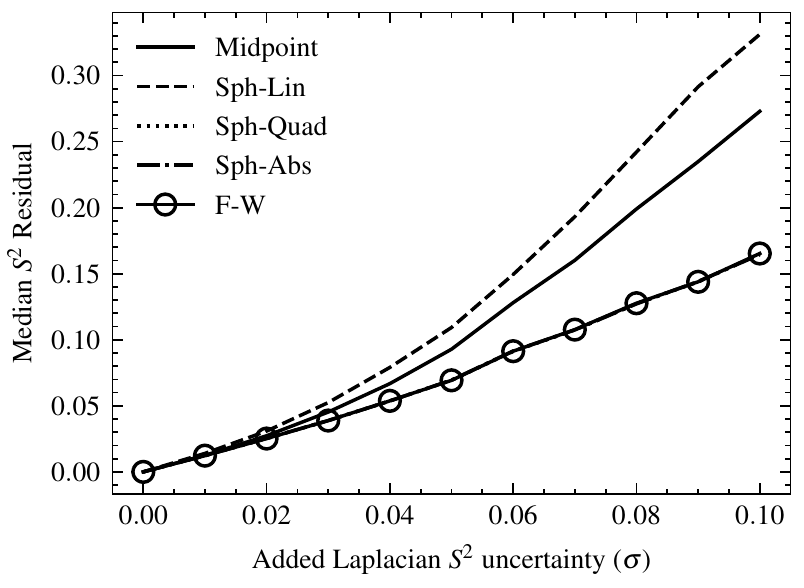}
    \end{subfigure} \hspace{1mm}
    \begin{subfigure}[b]{0.32\textwidth}
    \centering
        \includegraphics[height=4.3cm]{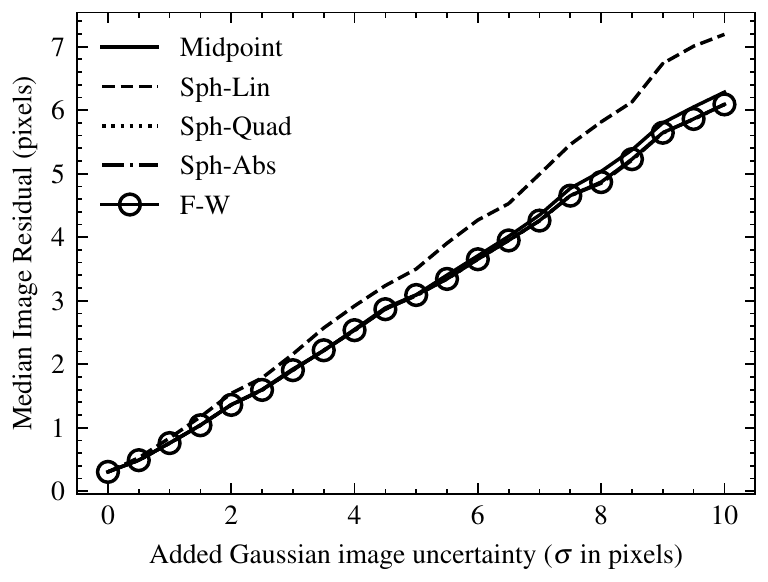}
    \end{subfigure} \hspace{1mm} 
    \vspace{-2mm}
    \caption{Residuals of the various spherical triangulation methods on the \textit{near points}, with both Gaussian (1st column) and Laplacian (2nd Column) uncertainty added on the projection sphere $S^2$. In all cases, \textbf{Sph-Quad}, \textbf{Sph-Abs} and \textbf{F-W} are almost identical. Regardless of whether Laplacian or Gaussian uncertainty is added, the methods perform similarly. Additionally, a fisheye camera is modelled, and Gaussian uncertainty is added on the modelled fisheye image (3rd column), in which case the \textbf{Midpoint} method performs marginally worse than the optimisation methods.}
    \label{fig:Spherical_Residual}
\end{figure*}

In Figure \ref{fig:Spherical_Residual}, the results for the spherical triangulation methods are presented, using the \textit{near points}. The sum-of-square optimisation approach (\textbf{Sph-Quad}) is optimal in the case of normal uncertainty. However, the sum-of-absolutes (\textbf{Sph-Abs}) optimisation is optimal for Laplacian uncertainty. For this reason, it makes sense check the performance of the triangulation method in the presence of both Gaussian and Laplacian uncertainty. Additionally, the set of points projected to the omnidirectional image $I^2$ has normal uncertainty applied on the image. These are unprojected to $S^2$ prior to spherical triangulation. In this case, none of the methods will be optimal, as the uncertainty model has been distorted by the process of raising the image points to the sphere.

\begin{figure*}
    \centering
    \begin{subfigure}[b]{0.32\textwidth}
        \includegraphics[height=4.3cm]{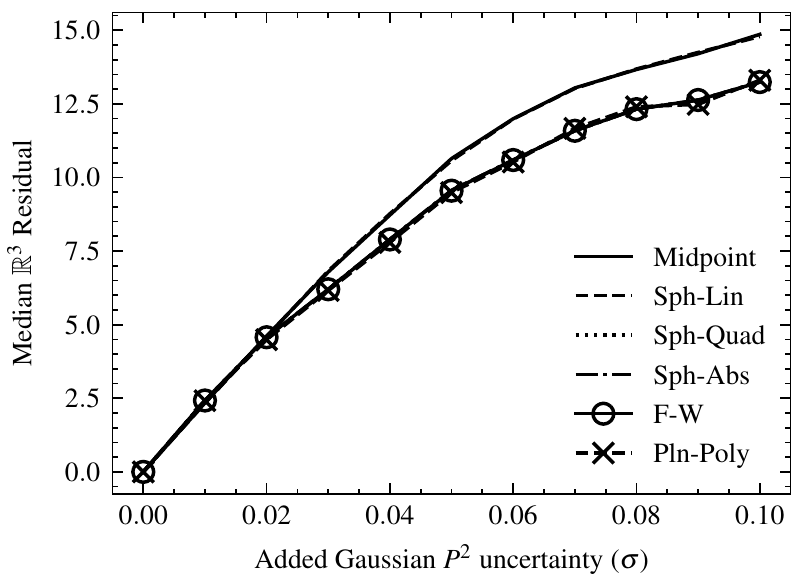}
    \end{subfigure}
    \begin{subfigure}[b]{0.32\textwidth}
        \includegraphics[height=4.3cm]{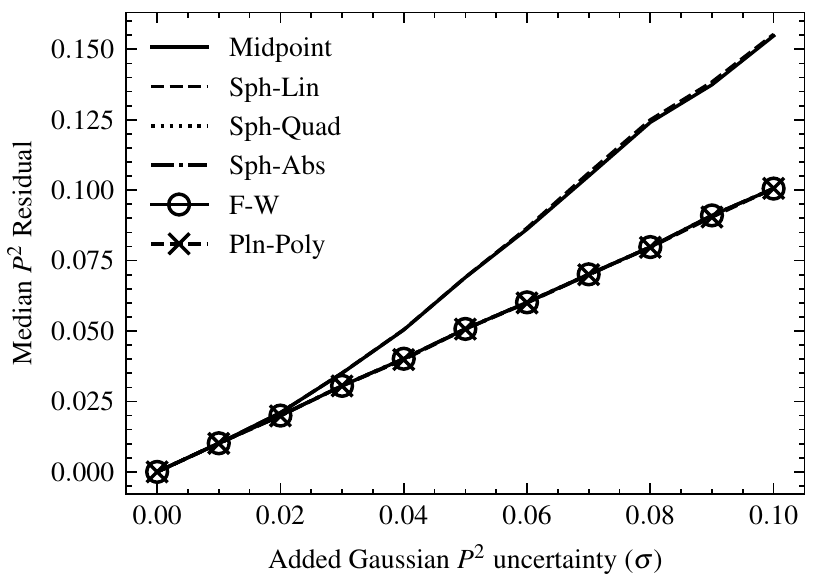}
    \end{subfigure}
    \begin{subfigure}[b]{0.32\textwidth}
        \includegraphics[height=4.3cm]{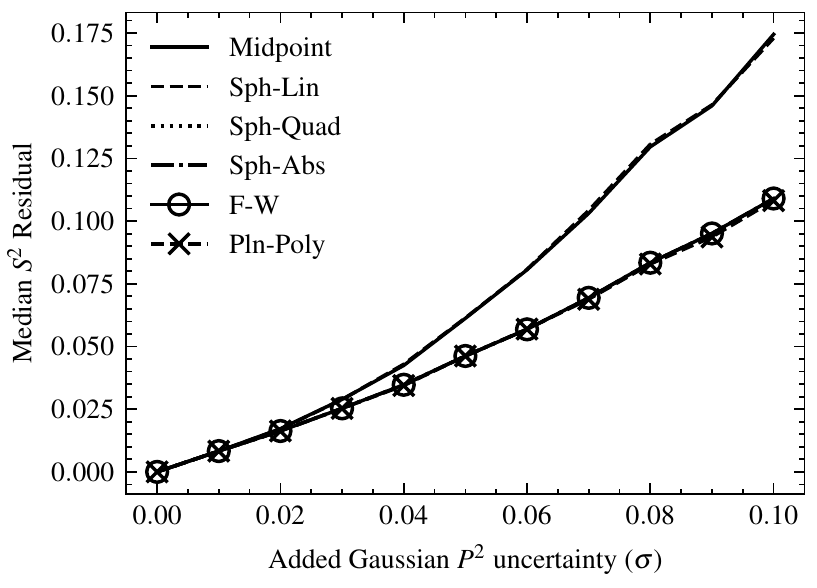}
    \end{subfigure}
    \vspace{-2mm}
    \caption{Residuals of the various triangulation methods (including the planar methods) for the \textit{far points}, with uncertainty added on the projective plane $P^2$. The \textbf{Sph-Quad}, \textbf{Sph-Abs}, \textbf{F-W} and \textbf{Pln-Poly} methods perform almost identically in all cases, with the \textbf{Midpoint} and \textbf{Sph-Lin} methods being significantly worse.}
    \label{fig:Linear_Residual}
    \vspace{-5mm}
\end{figure*}

Finally, it should be noticed that some of the steps in the \textbf{Sph-Quad} and \textbf{Sph-Abs} are simpler than the \textbf{Pln-Poly}, and as such should be computationally less expensive. It is therefore interesting to see how well the spherical methods work when applied to standard/narrow field-of-view cameras. In this case, the \textit{far points} are projected to the projective plane. Normal uncertainty is added to the projected features (i.i., on $P^2$). Results are given in Figure \ref{fig:Linear_Residual}. Even though the uncertainty is modelled on the $P^2$, residuals on $S^2$ and $\spce{R}^3$ are also presented for completeness.

\subsection{Evaluation on Real data}

Of course, with any real data sequences, the issue is always having a reliable ground truth. Calibration checkerboard sequences, thankfully, easily allow for an implicit ground truth without requiring an external reference. The sets of fisheye calibration images from \cite{Eichenseer2016Dataset, Urban2017} are used, along with two sets of narrow field-of-view images from \cite{matlabVision}. The checkerboard pattern is extracted using the OpenCV \texttt{findChessboardCorners} and \texttt{cornerSubPix} functions, and the ground truth is generated using \texttt{fisheye::calibrate} \cite{opencv_calibration}. The ground truth consists of the position and orientation of the cameras relative to the checkerboard pattern. The results are given in Table \ref{tab:real_values}. Again, the $L_2$-norm is used in both $P^2$ (reprojection), $S^2$ (reprojection), and $\spce{R}^3$ (3D) errors.


\begin{table*}[tbp]
\caption{Median $P^2$, $S^2$ and $\spce{R}^3$ errors for real calibration image datasets. \textbf{F-W} is not presented here, as it returns the same results as \textbf{Sph-Quad}.}
\vspace{-0.3cm}
\begin{center}
\begin{tabular}{c c c c c c c c c c c c c}
\hline
& \multicolumn{3}{c}{LMS Fisheye \cite{Eichenseer2016Dataset}} & \multicolumn{3}{c}{KIT IPF Fisheye \cite{Urban2017}} & \multicolumn{3}{c}{\cmulti{MATLAB GoPro \cite{matlabVision}\\(narrow, mild distortion)}} & \multicolumn{3}{c}{\cmulti{MATLAB Mono \cite{matlabVision}\\(narrow, no distortion)}} \\
 & \cmulti{$S^2$\\$\times 10^{-4}$} & \cmulti{$P^2$\\$\times 10^{-4}$} & \cmulti{$\spce{R}^3$\\$\times 10^{-2}$} & \cmulti{$S^2$\\$\times 10^{-4}$} & \cmulti{$P^2$\\$\times 10^{-4}$} & \cmulti{$\spce{R}^3$\\$\times 10^{-2}$} & \cmulti{$S^2$\\$\times 10^{-4}$} & \cmulti{$P^2$\\$\times 10^{-4}$} & \cmulti{$\spce{R}^3$\\$\times 10^{-3}$} & \cmulti{$S^2$\\$\times 10^{-5}$} & \cmulti{$P^2$\\$\times 10^{-4}$} & \cmulti{$\spce{R}^3$\\$\times 10^{-2}$} \\
\hline
Midpoint  & 2.010    & 4.238     & 1.214     & 4.058     & 8.336     & 1.347     & 1.982     & 3.643     & 6.281     & 5.247     & 1.311     & 1.797 \\
Sph-Lin   & 2.075    & 4.383     & 1.256     & 4.071     & 8.361     & 1.385     & 1.974     & 3.741     & 6.252     & 5.239     & 1.293     & 1.846 \\
Sph-Quad  & \bf1.672 & \bf3.786  & \bf1.007  & \bf3.784  & \bf8.123  & 1.249     & \bf1.785  & 3.366     & 6.004     & 5.068     & \bf1.201  & 1.734 \\
Sph-Abs   & 1.716    & 3.895     & 1.008     & 3.804     & 8.167     & \bf1.245  & 1.794     & 3.343     & \bf5.994  & \bf5.066  & 1.207     & 1.736 \\
Pln-Poly  & 1.986    & 4.128     & 1.148     & 4.080     & 8.413     & 1.366     & 1.809     & \bf3.340  & 5.999     & 5.068     & 1.202     & \bf1.733 \\
\hline
\end{tabular}
\label{tab:real_values}
\end{center}
\vspace{-0.2cm}
\end{table*}

\subsection{Runtime}

All methods have been implemented in Python. No specific optimisations have been performed, and Numpy functions have been used where appropriate (e.g., root finding, matrix inversion). Therefore, the results here should not be taken as an absolute measure of potential computation performance, but rather as a comparison between the approaches. 
Table \ref{tab:runtimes} shows the runtimes. Where \textit{opt. only} is noted, only the optimisation of the features on the surface is considered. Otherwise, the runtime includes the final triangulation. So that like-with-like can be compared, the \textbf{F-W} method with \textbf{Midpoint} method as final triangulation is also presented.

\begin{table}[tbp]
\caption{Runtime per triangulated points}
\vspace{-0.3cm}
\begin{center}
\begin{tabular}{c c}
\hline
Algorithm & Runtime ($\mu s$) \\
\hline
Midpoint & 11.7 \\
Sph-Lin & 114.6 \\
Sph-Quad (Midpoint) & 24.4 \\
Sph-Abs (Midpoint) & 28.4 \\
F-W & 865.9 \\
F-W (Midpoint) & 252.0 \\
Pln-Poly & 307.3 \\
\hline
Sph-Quad (Opt. only) & 14.7 \\
Sph-Abs (Opt. only) & 18.5 \\
F-W (Opt. Only) & 225.3 \\
Pln-Poly (Opt. Only) & 283.2 \\
\hline
\end{tabular}
\label{tab:runtimes}
\end{center}
\vspace{-5mm}
\end{table}

\subsection{Discussion}

Unsurprisingly, throughout the results, the \textbf{F-W} and \textbf{Sph-Quad} return the same results, tough \textbf{Sph-Quad} is computationally less expensive. Somewhat surprisingly, in the synthetic data, \textbf{Sph-Abs} is almost the same as \textbf{Sph-Quad} (Figure \ref{fig:Spherical_Residual}, left columns), regardless of the uncertainty model. There are divergences in the omnidirectional real data (Table \ref{tab:real_values}, left columns), where the \textbf{Sph-Quad} outperforms other methods, including the \textbf{Sph-Abs} in reprojection. Throughout, \textbf{Midpoint} and \textbf{Sph-Lin} perform worse than the optimisation methods, as would be expected.

In the synthetic \textit{far points} data (Figure \ref{fig:Linear_Residual}), the \textbf{Pln-Poly} method is only very slightly better than the spherical optimisation methods (\textbf{Sph-Quad} and \textbf{Sph-Abs}). In contrast, the \textbf{Midpoint} and the \textbf{Sph-Lin} methods perform quite poorly on this data. In the real data (Table \ref{tab:real_values}, right columns), there is little difference overall between the \textbf{Sph-Quad}, \textbf{Sph-Abs}, and \textbf{Pln-Poly} methods. This supports the earlier assertion that the spherical optimisation methods proposed here can be used just on narrow field-of-view wit reduced computational cost compared to \textbf{Pln-Poly}. However, it must be noted that the spherical methods are unsuitable for uncalibrated cameras, unlike \textbf{Pln-Poly}.

As expected, the \textbf{Midpoint} is computationally the fastest method. For the methods that optimise on the projection surface, the \textbf{Sph-Quad} and \textbf{Sph-Abs} methods are by a magnitude of order faster than the \textbf{F-W} and \textbf{Pln-Poly} methods.

\section{Conclusions}

\label{sec:conclusions}

In this paper, an efficient method for triangulation that optimises the reprojection error to a spherical surface is presented and is shown to boil down to the solution of a quadratic equation. The proposed direct spherical optimisation method, based on the sum-of-squares, returns the same result as that proposed by F\"{o}rstner and Wrobel \cite{Forstner2016book} within computational accuracy. A second method, based on the sum-of-absolutes, is also presented here (and also boils down to the solution of a quadratic), but performs slightly worse than the former in the real fisheye image datasets that were tested. Both presented methods presented are shown to be significantly computationally cheaper than existing methods. Finally, it is also shown that, when applied to calibrated ordinary field-of-view cameras, the results are essentially the same as the method of Hartley-Sturm \cite{hartley1997triangulation}. One must note, however, that the methods presented here are only defined for calibrated cameras, whereas that of Hartley-Sturm can be used in the uncalibrated case.

\ifCLASSOPTIONcaptionsoff
  \newpage
\fi

\bibliographystyle{IEEEtran}{}
\bibliography{bib/references}{}


\end{document}